\documentclass[12pt]{article}
\usepackage[english]{babel}

\usepackage{amsmath}
\usepackage{graphicx}
\usepackage[colorinlistoftodos]{todonotes}
\usepackage[utf8]{inputenc}
\usepackage{amsmath, amssymb, latexsym}
\usepackage{subcaption}
\usepackage[left=2cm,right=2cm,top=2cm,bottom=2cm]{geometry}
\usepackage{tikz}
\usetikzlibrary{decorations.pathreplacing}
\usetikzlibrary{fadings}

\begin{document}

\begin{titlepage}

\newcommand{\HRule}{\rule{\linewidth}{0.5mm}} 

\center 
 

\textsc{\LARGE Indian Institute of Technology Gandhinagar}\\[1.5cm] 


\HRule \\[0.4cm]
{ \huge \bfseries Deep Neural Networks for HDR imaging}\\[0.4cm] 
\HRule \\[1.5cm]
 

\begin{minipage}{0.4\textwidth}
\begin{flushleft} \large
\emph{Author:}\\
Kshiteej Sheth 
\end{flushleft}
\end{minipage}
~
\begin{minipage}{0.4\textwidth}
\begin{flushright} \large
\emph{Supervisor:} \\
Dr. Shanmuganathan Raman 
\end{flushright}
\end{minipage}\\[2cm]



{\large \today}\\[2cm] 


\includegraphics{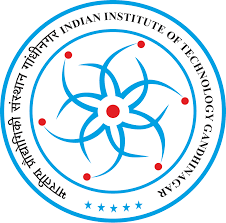}\\[1cm] 

-------------------------------------------------------------------

\vfill 

\end{titlepage}

\section*{Contents}
\renewcommand{\labelenumii}{\Roman{enumii}}
\begin{enumerate}
	\item Abstract
	\item Introduction 
    \item CNN Architecture
    \item Implementation details 
    \item Results
    \item References
\end{enumerate}


\section{Abstract}
We propose novel methods of solving two tasks using Convolutional Neural Networks, firstly the task of generating HDR map of a static scene using differently exposed LDR images of the scene captured using conventional cameras and secondly the task of finding an optimal tone mapping operator that would give a better score on the TMQI metric compared to the existing methods. We quantitatively show the performance of our networks and illustrate the cases where our networks performs good as well as bad.

\section{Introduction}

Natural scenes have a large range of intensity values(thus a large dynamic range) and conventional non-HDR cameras cannot capture this range in a single image. By controlling various factors, one of them being the exposure time of the shot we can capture a particular window in the total dynamic range of the scene. So we need multiple "Low dynamic range" images of the scene to get the complete information of the scene. Fig.1 illustrates an example. The internal processing pipeline of the camera is highly non linear i.e. the pixel intensity value at location $(i,j)$, $Z_{ij}$ is equal to  $f(E_{ij}\Delta t)$ where $\Delta t$ is the exposure time of the shot and $E_{ij}$ is the irradiance value at location $(i,j)$. $f(x)$ is known as the camera response function of that particular camera which is a non linear function. Given the values of $Z_{ij}$ for differently exposed images of the same scene we can get a robust,noise free estimate of the $E_{ij}$'s (methods like Debevec et. al. (1997) use a weighted average of the $f^{-1}(Z_{ij})/\Delta t$ to get a robust estimate of the corresponding $E_{ij}$. We use a deep neural network to estimate the function taking as input the LDR pixel intensities of 5 LDR images of a static scene to estimate the irradiance values, i.e. the HDR map of the scene. Fig 1. shows the camera acquisition pipeline in modern cameras and the non-linear transforms involved. 
\\ \\
We further conduct experiments of getting another similar convolutional neural network to approximate a tone mapping operator. Our training set includes HDR images of scenes and their corresponding tone mapped images generated by one of the Tone mapping operators provided in MATLAB's HDR-Toolbox (Banterle et al. 2011) which gives the highest value of the TMQI metric (Yeganeh et al. 2013). We try further experiments to improve the results, details of which are provided in the main report.
\\ \\
\includegraphics[width=\textwidth]{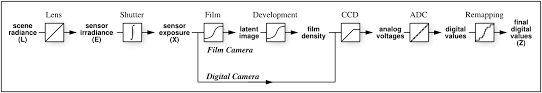}
\begin{center}
\footnotesize This is the end to end pipeline involved in digital image acquisition in modern digital cameras, which indicates that the mapping from the scene radiance to the pixel values at a spatial location in the image is non-linear.  
\end{center}

\section{CNN Architecture}
\label{sec:cnn}

\subsection{LDR2HDR network architecture}


We have a collected a dataset of 957 HDR images from the internet from the websites of various groups around the world working on HDR imaging and the camera response functions of the cameras with which those photos were taken. We use these to generate LDR images from the HDR's whose exposure time values are taken from a list of exposure times( a geometric progression with the first term 1, common ratio 4 and the last term equal to $4^9$. Our network takes as input a stack of 5 RGB LDR images of a particular scene, each having a different exposure. In the initial experiments we fixed the exposure of these LDR images to be [1,8,64,512,4096], and then we moved on to adaptive exposure based method in which we choose first the LDR image of a scene which gives the maximum value of entropy (entropy of a single channel image is defined as -$\sum p\log{p}$ where $p$ is the histogram count of a particular intensity value in the image and the sum is calculated for all the possible intensities in the image) and we take two images of the previous and next two intensity values. We have 3 networks, one for each of the R,G and B channels of the inputs, so We conducted many experiments with both the former and the latter case approach but we were able to obtain plausible results only in the first case where the exposure times were fixed. The graphs of training error vs. epochs for 3 different models which turned out to be the best after testing many different sets of hyperparameters are shown below for the case where the exposure times are kept constant. The final test error that we get for the best model is . We conducted experiments in models with dropout in each layer with $p=0.4$ in order to improve generalization. We also also added spatial batch normalization just before passing the activations of each layer through the ReLU non linearity. Batch size was kept 40 (decided by the memory limitations of the GPU). BatchNorm strictly improved the results as the same training error was attained in less number of epochs. The architecture of the network is illustrated in the Fig \ref{fig:tone_cnn}

\subsection{HDR2ToneMap network architecture}

We first create a dataset using the existing 957 HDR images. We then use the tone mapping operators provided in the HDR-toolbox by Francesco Banterle et al. and use them to create different tone maps of each HDR and run the TMQI metric on each of the tone maps and choose the one which gives the highest TMQI score. Some are local and some are global tone mapping operators, so our approach is not fully justified. We then train a convolutional neural network whose input is the HDR image and its corresponding truth is the best Tone map corresponding to the TMQI metric. We use 3x3 conv in the first layer followed by 1x1 convs in the subsequent layers. We get this intuition from the works of Mertens et al. whose method most of the time gave the best TMQI score. In their work the final intensity value at location $(x,y)$ depends only on the 3x3 neighborhood of radiance value in its corresponding HDR image at location $(x,y)$. Further study of the other tone mapping works is required in order to improve the architecture after the pilot testing that we have done in the course of the summer. After preliminary results it was observed that the network was not able to deal with high frequency inputs simultaneously with low frequency ones so in order to tackle that problem we first convert both the input and output pairs to Lab space, apply a Bilateral Filter to the $L$ channel, create a new channel $L_{original} - L_{filtered}$ and train 4 networks each for these new channels as well as for a and b channels. We obtain better results for this method. In order to obtain a good estimate of the hyperparameters of the network, we test out several values of then by training their corresponding architectures for 2 epochs and observing the validation error. Due to computational constraints, this could not be afforded for more epochs and only 4 sets of hyperparameters could be tested.

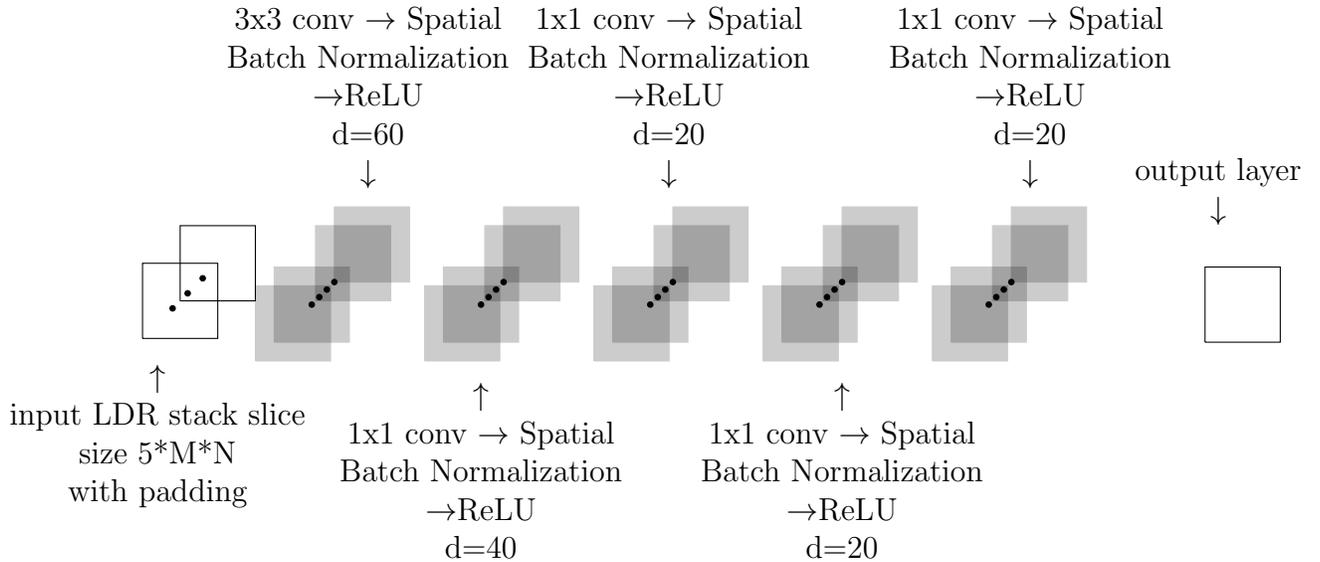
\begin{figure}[h!]
	\centering
	\begin{tikzpicture}
		\node at (0.2,-1){\begin{tabular}{c}$\uparrow$\\input LDR stack slice \\size 5*M*N \\with padding\end{tabular}};
		
        \draw (0.0,0.3) -- (1.0,0.3) -- (1.0,1.3) -- (0.0,1.3) -- (0.0,0.3);
        \draw[fill=black] (0.4,0.7) circle (0.2ex);
        \draw[fill=black] (0.6,0.9) circle (0.2ex);
        \draw[fill=black] (0.8,1.1) circle (0.2ex);
        \draw (0.5,0.8) -- (1.5,0.8) -- (1.5,1.8) -- (0.5,1.8) -- (0.5,0.8);
		\node at (3,3.5){\begin{tabular}{c}3x3 conv $\rightarrow$ Spatial\\Batch Normalization\\$\rightarrow$ReLU\\d=60\\$\downarrow$\end{tabular}};
		
		\draw[fill=black,opacity=0.2,draw=black] (2.55,1.05) -- (3.55,1.05) -- (3.55,2.05) -- (2.55,2.05) -- (2.55,1.05);
		\draw[fill=black,opacity=0.2,draw=black] (2.3,0.8) -- (3.3,0.8) -- (3.3,1.8) -- (2.3,1.8) -- (2.3,0.8);
        \draw[fill=black] (2.25,0.75) circle (0.2ex);
        \draw[fill=black] (2.35,0.85) circle (0.2ex);
        \draw[fill=black] (2.45,0.95) circle (0.2ex);
        \draw[fill=black] (2.55,1.05) circle (0.2ex);

		\draw[fill=black,opacity=0.2,draw=black] (1.75,0.25) -- (2.75,0.25) -- (2.75,1.25) -- (1.75,1.25) -- (1.75,0.25);
		\draw[fill=black,opacity=0.2,draw=black] (1.5,0) -- (2.5,0) -- (2.5,1) -- (1.5,1) -- (1.5,0);
		
		\node at (4.5,-1.5){\begin{tabular}{c}$\uparrow$\\1x1 conv $\rightarrow$ Spatial\\Batch Normalization\\$\rightarrow$ReLU\\d=40\end{tabular}};
		
		\draw[fill=black,opacity=0.2,draw=black] (2.55+2.25,1.05) -- (3.55+2.25,1.05) -- (3.55+2.25,2.05) -- (2.55+2.25,2.05) -- (2.55+2.25,1.05);
		\draw[fill=black,opacity=0.2,draw=black] (2.3+2.25,0.8) -- (3.3+2.25,0.8) -- (3.3+2.25,1.8) -- (2.3+2.25,1.8) -- (2.3+2.25,0.8);
        \draw[fill=black] (2.25+2.25,0.75) circle (0.2ex);
        \draw[fill=black] (2.35+2.25,0.85) circle (0.2ex);
        \draw[fill=black] (2.45+2.25,0.95) circle (0.2ex);
        \draw[fill=black] (2.55+2.25,1.05) circle (0.2ex);

		\draw[fill=black,opacity=0.2,draw=black] (1.75+2.25,0.25) -- (2.75+2.25,0.25) -- (2.75+2.25,1.25) -- (1.75+2.25,1.25) -- (1.75+2.25,0.25);
		\draw[fill=black,opacity=0.2,draw=black] (3.75,0) -- (4.75,0) -- (4.75,1) -- (3.75,1) -- (3.75,0);
		
		\node at (7,3.5){\begin{tabular}{c}1x1 conv $\rightarrow$ Spatial\\Batch Normalization\\$\rightarrow$ReLU\\d=20\\$\downarrow$\end{tabular}};
		
		\draw[fill=black,opacity=0.2,draw=black] (2.55+4.5,1.05) -- (3.55+4.5,1.05) -- (3.55+4.5,2.05) -- (2.55+4.5,2.05) -- (2.55+4.5,1.05);
		\draw[fill=black,opacity=0.2,draw=black] (2.3+4.5,0.8) -- (3.3+4.5,0.8) -- (3.3+4.5,1.8) -- (2.3+4.5,1.8) -- (2.3+4.5,0.8);
        \draw[fill=black] (2.25+4.5,0.75) circle (0.2ex);
        \draw[fill=black] (2.35+4.5,0.85) circle (0.2ex);
        \draw[fill=black] (2.45+4.5,0.95) circle (0.2ex);
        \draw[fill=black] (2.55+4.5,1.05) circle (0.2ex);

		\draw[fill=black,opacity=0.2,draw=black] (1.75+4.5,0.25) -- (2.75+4.5,0.25) -- (2.75+4.5,1.25) -- (1.75+4.5,1.25) -- (1.75+4.5,0.25);
		\draw[fill=black,opacity=0.2,draw=black] (1.5+4.5,0) -- (2.5+4.5,0) -- (2.5+4.5,1) -- (1.5+4.5,1) -- (1.5+4.5,0);
		
		\node at (9.3,-1.5){\begin{tabular}{c}$\uparrow$\\1x1 conv $\rightarrow$ Spatial\\Batch Normalization\\$\rightarrow$ReLU\\d=20\end{tabular}};

		\draw[fill=black,opacity=0.2,draw=black] (2.55+4.5+2.25,1.05) -- (3.55+4.5+2.25,1.05) -- (3.55+4.5+2.25,2.05) -- (2.55+4.5+2.25,2.05) -- (2.55+4.5+2.25,1.05);
		\draw[fill=black,opacity=0.2,draw=black] (2.3+4.5+2.25,0.8) -- (3.3+4.5+2.25,0.8) -- (3.3+4.5+2.25,1.8) -- (2.3+4.5+2.25,1.8) -- (2.3+4.5+2.25,0.8);
        \draw[fill=black] (2.25+4.5+2.25,0.75) circle (0.2ex);
        \draw[fill=black] (2.35+4.5+2.25,0.85) circle (0.2ex);
        \draw[fill=black] (2.45+4.5+2.25,0.95) circle (0.2ex);
        \draw[fill=black] (2.55+4.5+2.25,1.05) circle (0.2ex);

		\draw[fill=black,opacity=0.2,draw=black] (1.75+4.5+2.25,0.25) -- (2.75+4.5+2.25,0.25) -- (2.75+4.5+2.25,1.25) -- (1.75+4.5+2.25,1.25) -- (1.75+4.5+2.25,0.25);
		\draw[fill=black,opacity=0.2,draw=black] (1.5+4.5+2.25,0) -- (2.5+4.5+2.25,0) -- (2.5+4.5+2.25,1) -- (1.5+4.5+2.25,1) -- (1.5+4.5+2.25,0);
        
		\node at (11.8,3.5){\begin{tabular}{c}1x1 conv $\rightarrow$ Spatial\\Batch Normalization\\$\rightarrow$ReLU\\d=20\\$\downarrow$\end{tabular}};

		\draw[fill=black,opacity=0.2,draw=black] (2.55+4.5+2.25+2.25,1.05) -- (3.55+4.5+2.25+2.25,1.05) -- (3.55+4.5+2.25+2.25,2.05) -- (2.55+4.5+2.25+2.25,2.05) -- (2.55+4.5+2.25+2.25,1.05);
		\draw[fill=black,opacity=0.2,draw=black] (2.3+4.5+2.25+2.25,0.8) -- (3.3+4.5+2.25+2.25,0.8) -- (3.3+4.5+2.25+2.25,1.8) -- (2.3+4.5+2.25+2.25,1.8) -- (2.3+4.5+2.25+2.25,0.8);
        \draw[fill=black] (2.25+4.5+2.25+2.25,0.75) circle (0.2ex);
        \draw[fill=black] (2.35+4.5+2.25+2.25,0.85) circle (0.2ex);
        \draw[fill=black] (2.45+4.5+2.25+2.25,0.95) circle (0.2ex);
        \draw[fill=black] (2.55+4.5+2.25+2.25,1.05) circle (0.2ex);

		\draw[fill=black,opacity=0.2,draw=black] (1.75+4.5+2.25+2.25,0.25) -- (2.75+4.5+2.25+2.25,0.25) -- (2.75+4.5+2.25+2.25,1.25) -- (1.75+4.5+2.25+2.25,1.25) -- (1.75+4.5+2.25+2.25,0.25);
		\draw[fill=black,opacity=0.2,draw=black] (1.5+4.5+2.25+2.25,0) -- (2.5+4.5+2.25+2.25,0) -- (2.5+4.5+2.25+2.25,1) -- (1.5+4.5+2.25+2.25,1) -- (1.5+4.5+2.25+2.25,0);
        
		\node at (14.3,2.25){\begin{tabular}{c}output layer\\$\downarrow$\end{tabular}};
				\draw (1.75+4.5+2.25+2.25+2.25+1.125,0.25) -- (2.75+4.5+2.25+2.25+2.25+1.125,0.25) -- (2.75+4.5+2.25+2.25+2.25+1.125,1.25) -- (1.75+4.5+2.25+2.25+2.25+1.125,1.25) -- (1.75+4.5+2.25+2.25+2.25+1.125,0.25);

	\end{tikzpicture}
	\caption[Architecture of a traditional convolutional neural network.]{This is the network architecture which is used to generate the HDR image from the LDR stack. There are 3 networks for each of the
    R,G and B channels.}
	\label{fig:tone_cnn}
\end{figure}
\begin{figure}[h!]
	\centering
	\begin{tikzpicture}
		\node at (0.2,-1){\begin{tabular}{c}$\uparrow$\\input HDR slice \\size 1*M*N \\with padding\end{tabular}};
		
        \draw (0.0,0.3) -- (1.0,0.3) -- (1.0,1.3) -- (0.0,1.3) -- (0.0,0.3);
        
		\node at (3,3.5){\begin{tabular}{c}3x3 conv $\rightarrow$ Spatial\\Batch Normalization\\$\rightarrow$ReLU\\d=100\\$\downarrow$\end{tabular}};
		
		\draw[fill=black,opacity=0.2,draw=black] (2.55,1.05) -- (3.55,1.05) -- (3.55,2.05) -- (2.55,2.05) -- (2.55,1.05);
		\draw[fill=black,opacity=0.2,draw=black] (2.3,0.8) -- (3.3,0.8) -- (3.3,1.8) -- (2.3,1.8) -- (2.3,0.8);
        \draw[fill=black] (2.25,0.75) circle (0.2ex);
        \draw[fill=black] (2.35,0.85) circle (0.2ex);
        \draw[fill=black] (2.45,0.95) circle (0.2ex);
        \draw[fill=black] (2.55,1.05) circle (0.2ex);

		\draw[fill=black,opacity=0.2,draw=black] (1.75,0.25) -- (2.75,0.25) -- (2.75,1.25) -- (1.75,1.25) -- (1.75,0.25);
		\draw[fill=black,opacity=0.2,draw=black] (1.5,0) -- (2.5,0) -- (2.5,1) -- (1.5,1) -- (1.5,0);
		
		\node at (4.5,-1.5){\begin{tabular}{c}$\uparrow$\\1x1 conv $\rightarrow$ Spatial\\Batch Normalization\\$\rightarrow$ReLU\\d=80\end{tabular}};
		
		\draw[fill=black,opacity=0.2,draw=black] (2.55+2.25,1.05) -- (3.55+2.25,1.05) -- (3.55+2.25,2.05) -- (2.55+2.25,2.05) -- (2.55+2.25,1.05);
		\draw[fill=black,opacity=0.2,draw=black] (2.3+2.25,0.8) -- (3.3+2.25,0.8) -- (3.3+2.25,1.8) -- (2.3+2.25,1.8) -- (2.3+2.25,0.8);
        \draw[fill=black] (2.25+2.25,0.75) circle (0.2ex);
        \draw[fill=black] (2.35+2.25,0.85) circle (0.2ex);
        \draw[fill=black] (2.45+2.25,0.95) circle (0.2ex);
        \draw[fill=black] (2.55+2.25,1.05) circle (0.2ex);

		\draw[fill=black,opacity=0.2,draw=black] (1.75+2.25,0.25) -- (2.75+2.25,0.25) -- (2.75+2.25,1.25) -- (1.75+2.25,1.25) -- (1.75+2.25,0.25);
		\draw[fill=black,opacity=0.2,draw=black] (3.75,0) -- (4.75,0) -- (4.75,1) -- (3.75,1) -- (3.75,0);
		
		\node at (7,3.5){\begin{tabular}{c}1x1 conv $\rightarrow$ Spatial\\Batch Normalization\\$\rightarrow$ReLU\\d=50\\$\downarrow$\end{tabular}};
		
		\draw[fill=black,opacity=0.2,draw=black] (2.55+4.5,1.05) -- (3.55+4.5,1.05) -- (3.55+4.5,2.05) -- (2.55+4.5,2.05) -- (2.55+4.5,1.05);
		\draw[fill=black,opacity=0.2,draw=black] (2.3+4.5,0.8) -- (3.3+4.5,0.8) -- (3.3+4.5,1.8) -- (2.3+4.5,1.8) -- (2.3+4.5,0.8);
        \draw[fill=black] (2.25+4.5,0.75) circle (0.2ex);
        \draw[fill=black] (2.35+4.5,0.85) circle (0.2ex);
        \draw[fill=black] (2.45+4.5,0.95) circle (0.2ex);
        \draw[fill=black] (2.55+4.5,1.05) circle (0.2ex);

		\draw[fill=black,opacity=0.2,draw=black] (1.75+4.5,0.25) -- (2.75+4.5,0.25) -- (2.75+4.5,1.25) -- (1.75+4.5,1.25) -- (1.75+4.5,0.25);
		\draw[fill=black,opacity=0.2,draw=black] (1.5+4.5,0) -- (2.5+4.5,0) -- (2.5+4.5,1) -- (1.5+4.5,1) -- (1.5+4.5,0);
		
		\node at (9.3,-1.5){\begin{tabular}{c}$\uparrow$\\1x1 conv $\rightarrow$ Spatial\\Batch Normalization\\$\rightarrow$ReLU\\d=10\end{tabular}};

		\draw[fill=black,opacity=0.2,draw=black] (2.55+4.5+2.25,1.05) -- (3.55+4.5+2.25,1.05) -- (3.55+4.5+2.25,2.05) -- (2.55+4.5+2.25,2.05) -- (2.55+4.5+2.25,1.05);
		\draw[fill=black,opacity=0.2,draw=black] (2.3+4.5+2.25,0.8) -- (3.3+4.5+2.25,0.8) -- (3.3+4.5+2.25,1.8) -- (2.3+4.5+2.25,1.8) -- (2.3+4.5+2.25,0.8);
        \draw[fill=black] (2.25+4.5+2.25,0.75) circle (0.2ex);
        \draw[fill=black] (2.35+4.5+2.25,0.85) circle (0.2ex);
        \draw[fill=black] (2.45+4.5+2.25,0.95) circle (0.2ex);
        \draw[fill=black] (2.55+4.5+2.25,1.05) circle (0.2ex);

		\draw[fill=black,opacity=0.2,draw=black] (1.75+4.5+2.25,0.25) -- (2.75+4.5+2.25,0.25) -- (2.75+4.5+2.25,1.25) -- (1.75+4.5+2.25,1.25) -- (1.75+4.5+2.25,0.25);
		\draw[fill=black,opacity=0.2,draw=black] (1.5+4.5+2.25,0) -- (2.5+4.5+2.25,0) -- (2.5+4.5+2.25,1) -- (1.5+4.5+2.25,1) -- (1.5+4.5+2.25,0);
		
		\node at (11.3,2.25){\begin{tabular}{c}output layer\\$\downarrow$\end{tabular}};
				\draw (1.75+4.5+2.25+2.25,0.25) -- (2.75+4.5+2.25+2.25,0.25) -- (2.75+4.5+2.25+2.25,1.25) -- (1.75+4.5+2.25+2.25,1.25) -- (1.75+4.5+2.25+2.25,0.25);

	\end{tikzpicture}
	\caption[Architecture of a traditional convolutional neural network.]{This is the network architecture which is used to approximate the tone mapping operator, there are four networks, one for the bilaterally filtered L channel, Original L minus the filtered L channel, a and b channel of the of the HDR image}
	\label{fig:tone_cnn}
\end{figure}

\section{Implementation Details}

For all the data processing tasks we use MATLAB and for implementing and testing our neural networks we use the Torch framework in the Lua scripting language, and most of the models are trained on a single NVIDIA GeForce GT 730 graphics processor, although for a brief amount of time during which we had access to a HPC node which had 2 GPU's, a Tesla K20C and a Titan X, we did multi-GPU training of our models using the following algorithm - 
\begin{itemize}
	\item Have the same network on the 2 GPU's at the beginning of every iteration in an epoch.
    \item Independently processing two different batches on the two GPU's and then copying over the accumulated gradients in the backward pass on one of the GPU's to the other, adding them to the accumulated gradients on the other GPU during the backward pass on it .
    \item Updating the parameters of the model on the GPU to which the gradients were copied.
    \item Process the next set of batches
\end{itemize}
One drawback with this approach was that the inter GPU communication overhead outweighed the almost 2X gain time in the actual training of the networks (the time required to the forward-backward pass).
During our other experiments, in order to save time in loading of the data, we implemented a multi-threaded approach to load our mini-batch. \\
Another important thing to note is that we were not able to process even a single input example of dimensions 15 X M X N as our images were of quite high resolution and during a forward pass of the network since every individual module in Torch caches its local output, the GPU's memory didn't turn out to be sufficient, so we broke each image into patches of 64 X 64 and after that we were able to keep the minibatch size to be 40 without overloading the GPU's memory. Due to computing power issues we were not able to test our models that efficiently. The code will shortly be made available at my github repository.

\section{Results}
\subsection{LDR2HDR Results}
In the results we present the graph of the training error vs no. of epochs for the best three models(Fig 3.). The test error for the best model is 0.09345. Visual results are shown below. It is clear that further experiments with validating the hyperparameters are required to find the optimal architecture for the task. The network is clearly able to generate plausible results for some of the colors but not all(Fig. 3 and 4). Also it is clear that the network is able to generate outputs without under/over saturation of regions that have high and low radiance values in the same image which hence proves that the dynamic range of the output is quite high.
\begin{figure}[bp!]
\includegraphics[width = \textwidth]{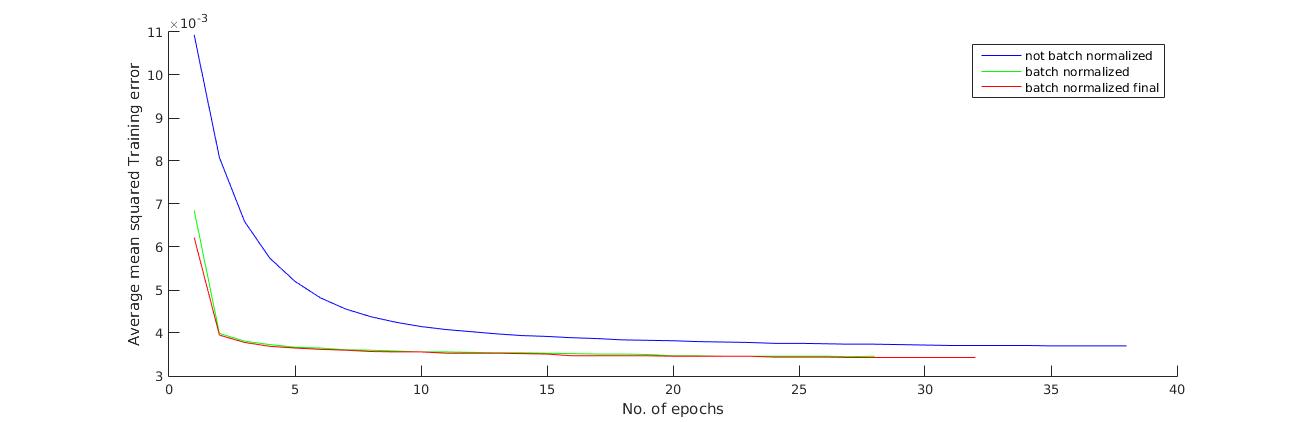}
\caption{The training error plot for the three best architectures, blue and green share the same architecture, whereas the final one,red, is one more no. of activations in the first layer. The architecture of the final network has been illustrated in Fig 1.)}
\end{figure}
\begin{figure}[bp!]
\begin{subfigure}{0.5\textwidth}
\includegraphics[scale = 0.62]{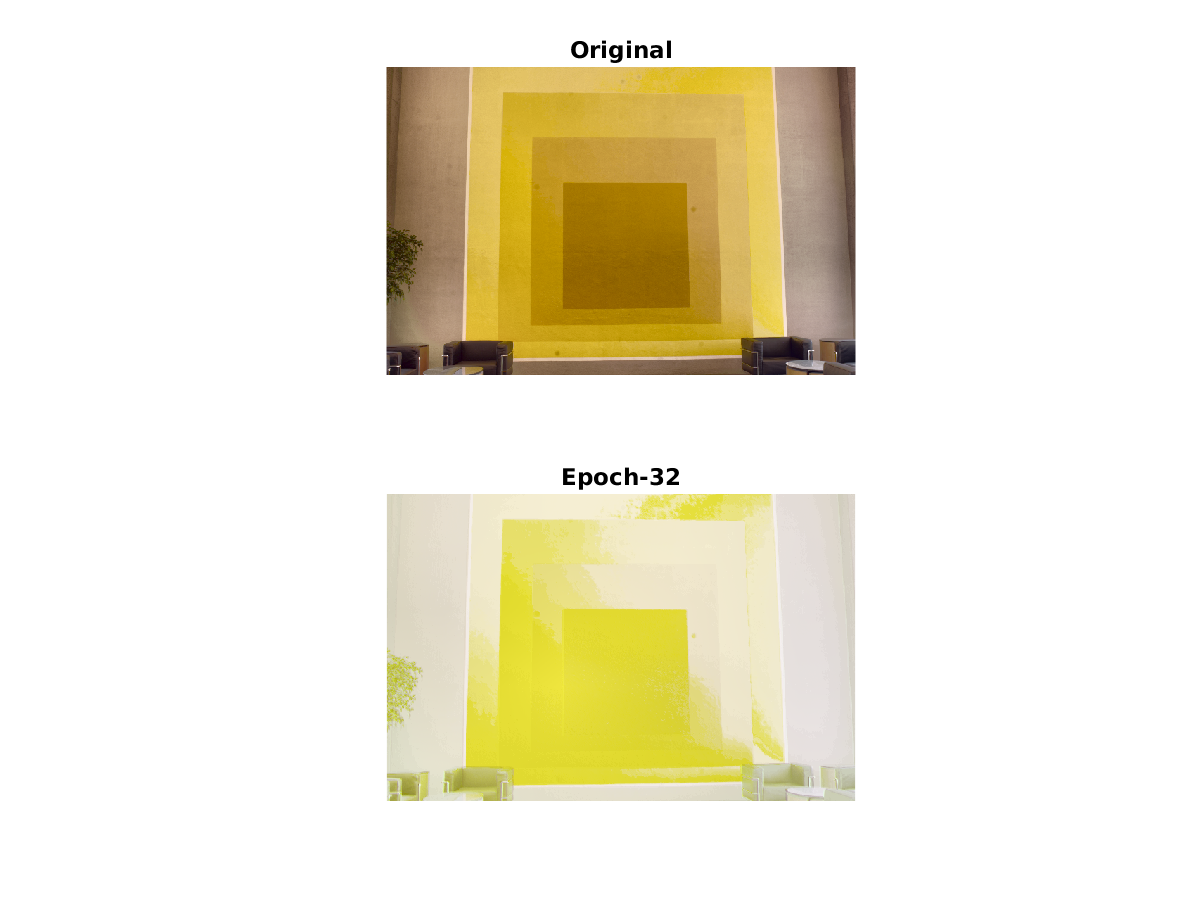}
\end{subfigure}
\begin{subfigure}{0.5\textwidth}
\includegraphics[scale = 0.62]{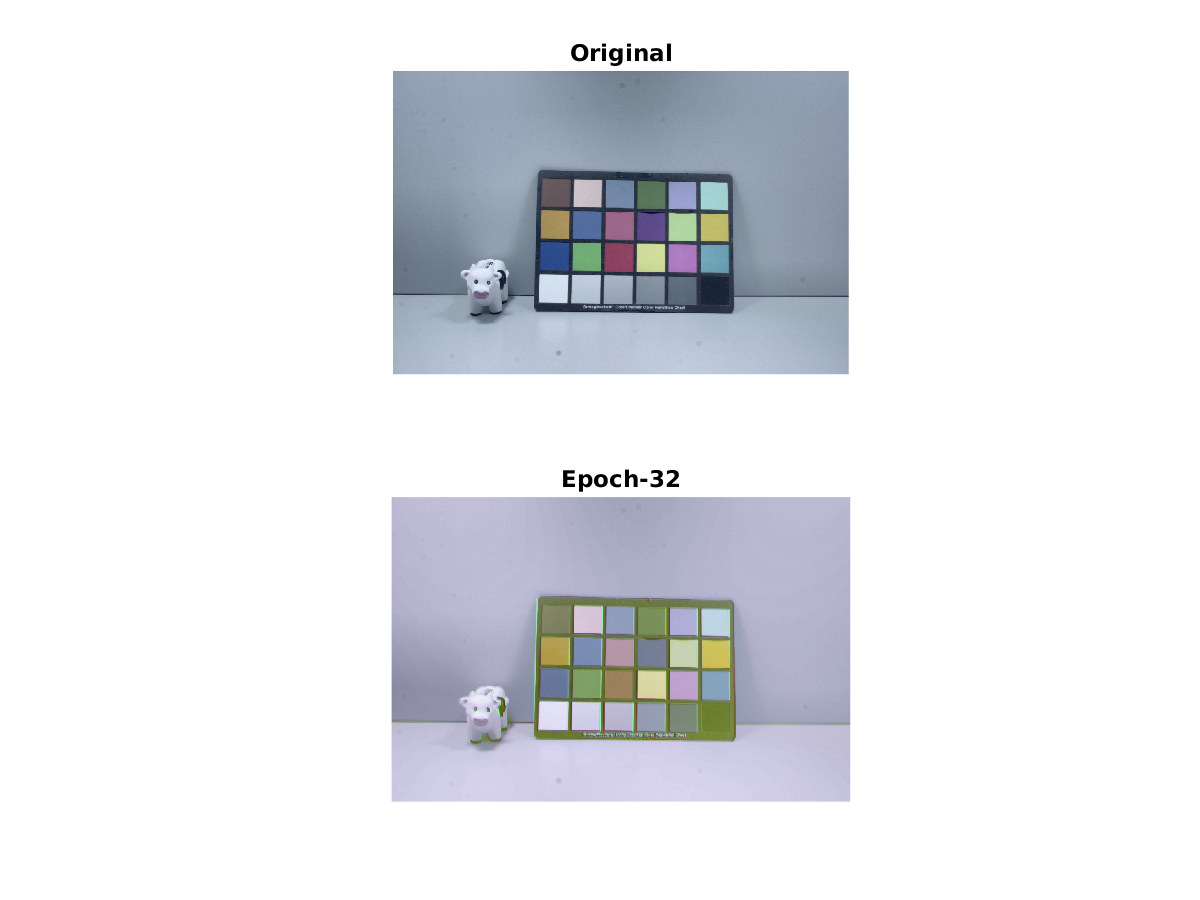}
\end{subfigure}
\caption{Some examples from test set where the LDR2HDR network gives plausible results.}
\end{figure}
\begin{figure}[bp!]
\begin{subfigure}{0.5\textwidth}
\includegraphics[scale = 0.62]{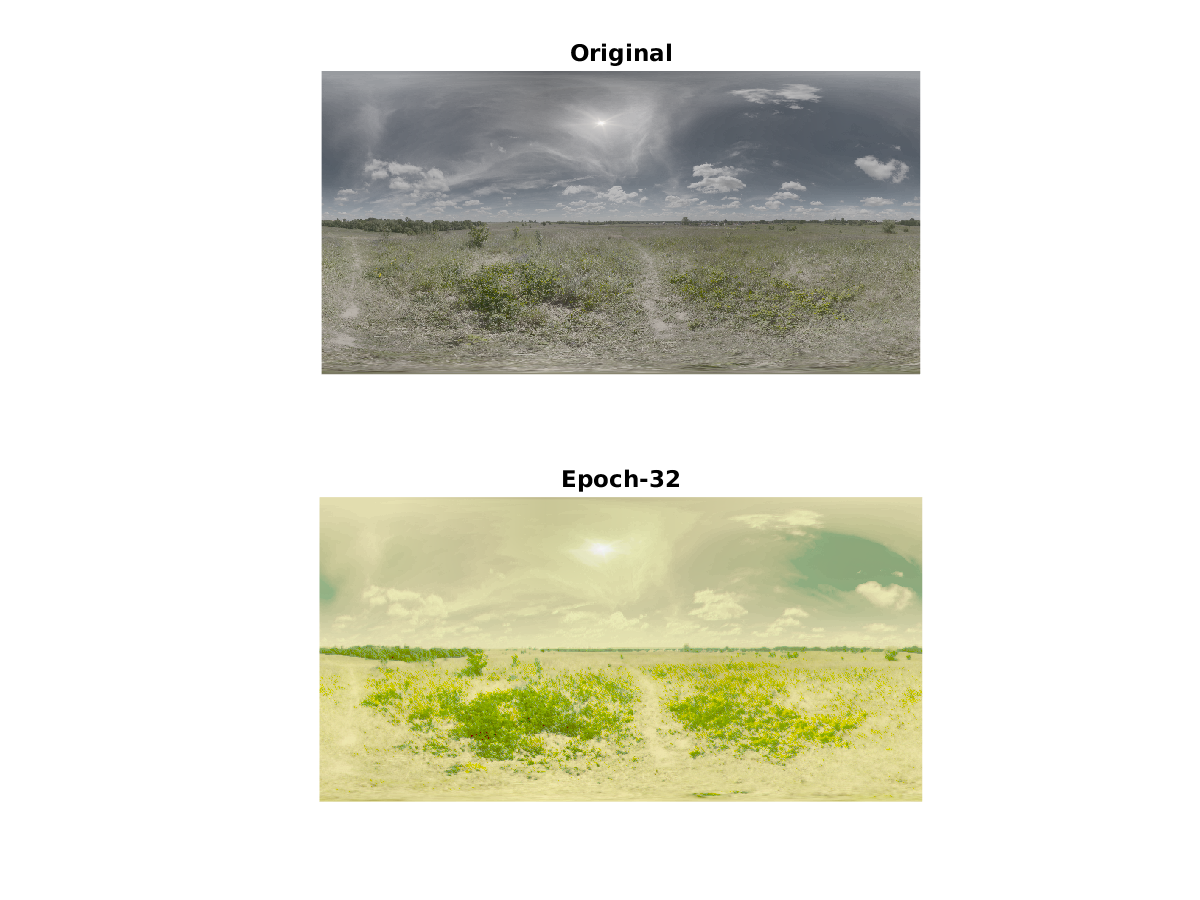}
\end{subfigure}
\begin{subfigure}{0.5\textwidth}
\includegraphics[scale = 0.62]{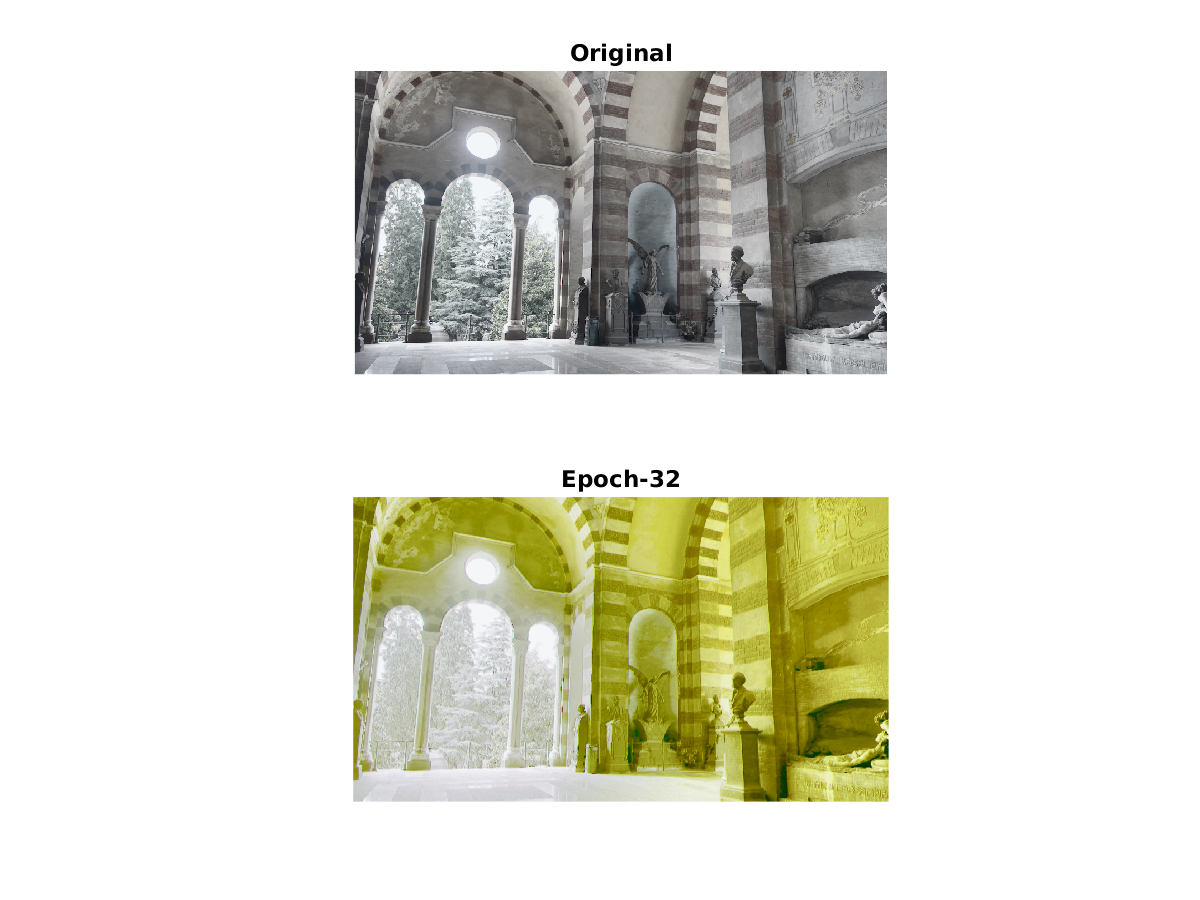}
\end{subfigure}
\caption{Some examples from test set where the LDR2HDR network does not give plausible results.}
\end{figure}
\begin{figure}[bp!]
\includegraphics[width = \textwidth]{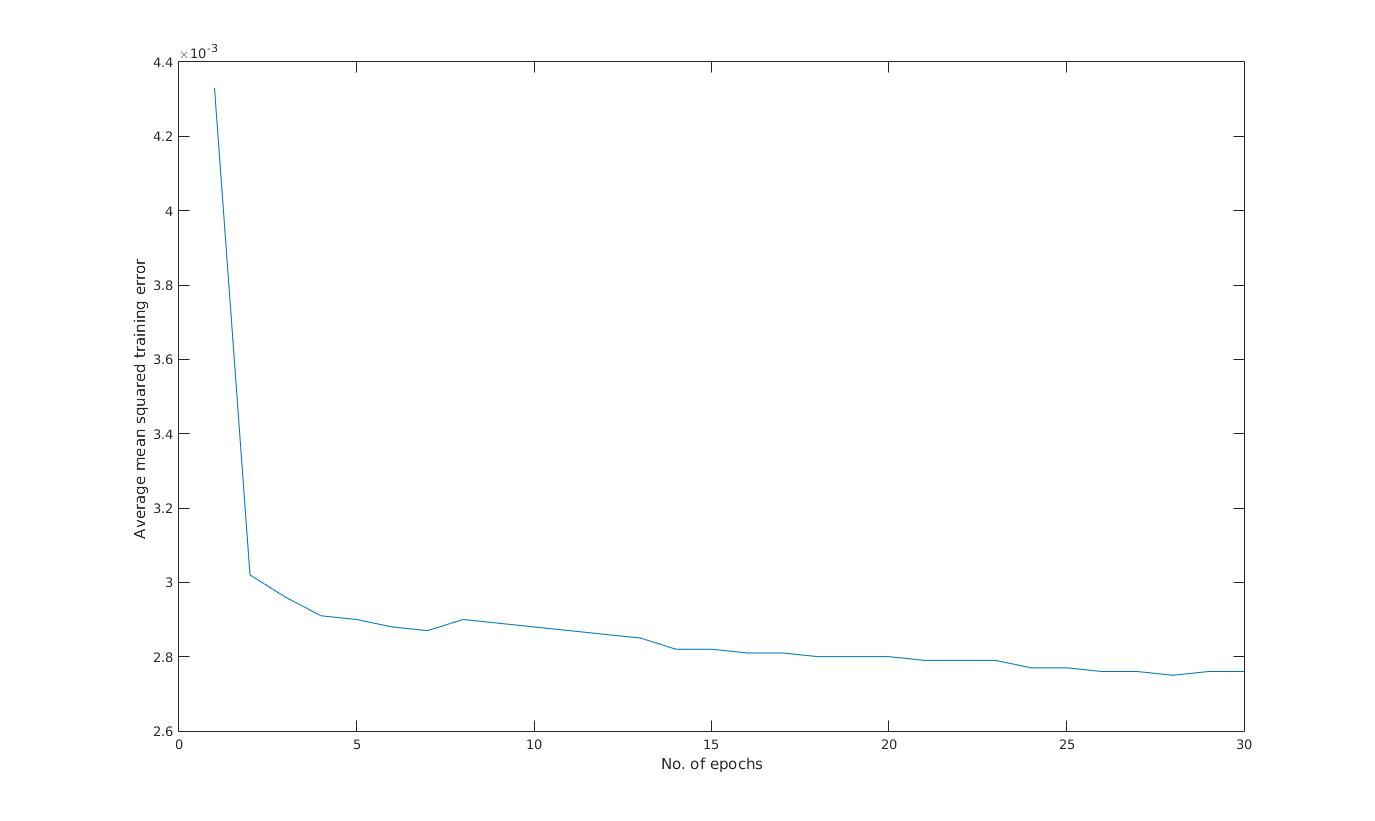}
\caption{The training error plot for the final tone mapping network whose architecture has been illustrated in Fig 2.}
\end{figure}
\begin{figure}
\begin{subfigure}{0.5\textwidth}
\includegraphics[scale = 0.5]{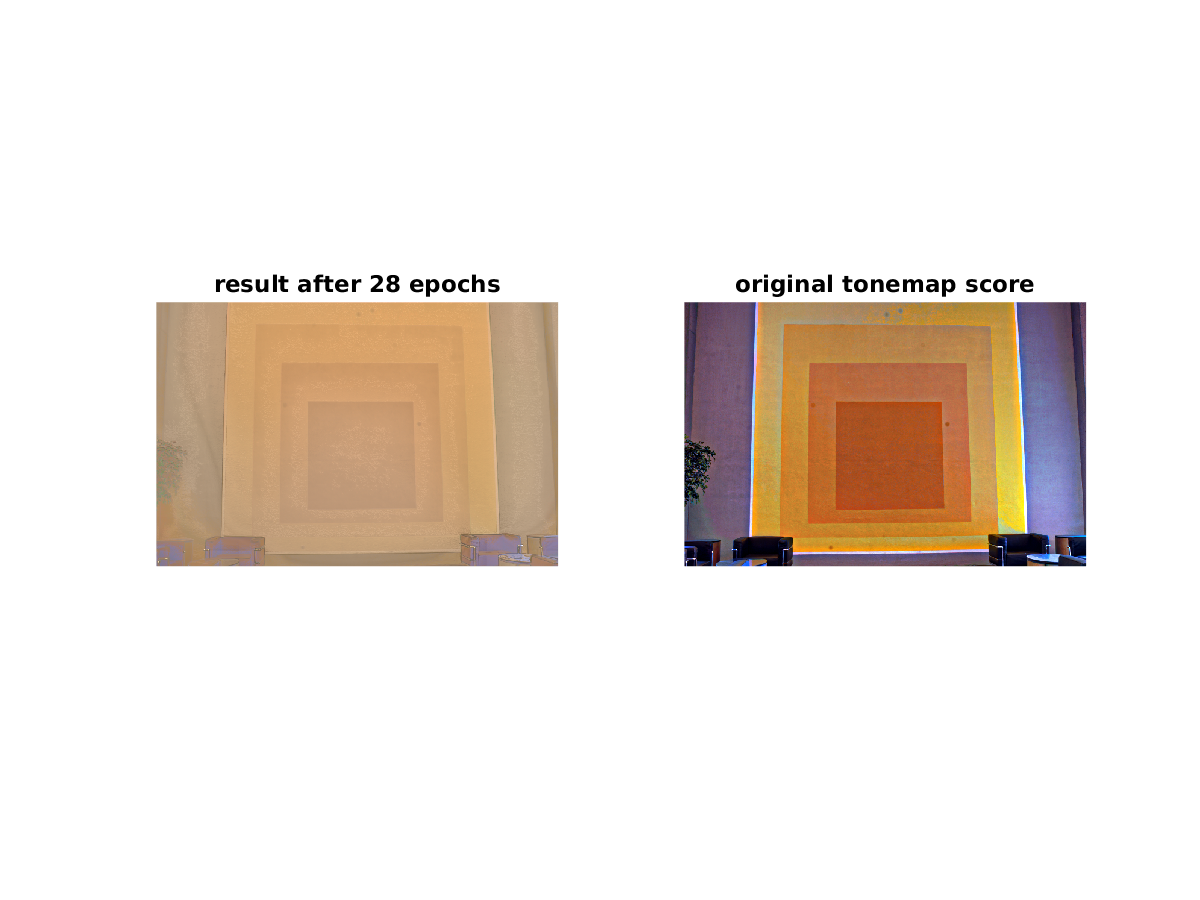}
\end{subfigure}
\begin{subfigure}{0.5\textwidth}
\includegraphics[scale = 0.5]{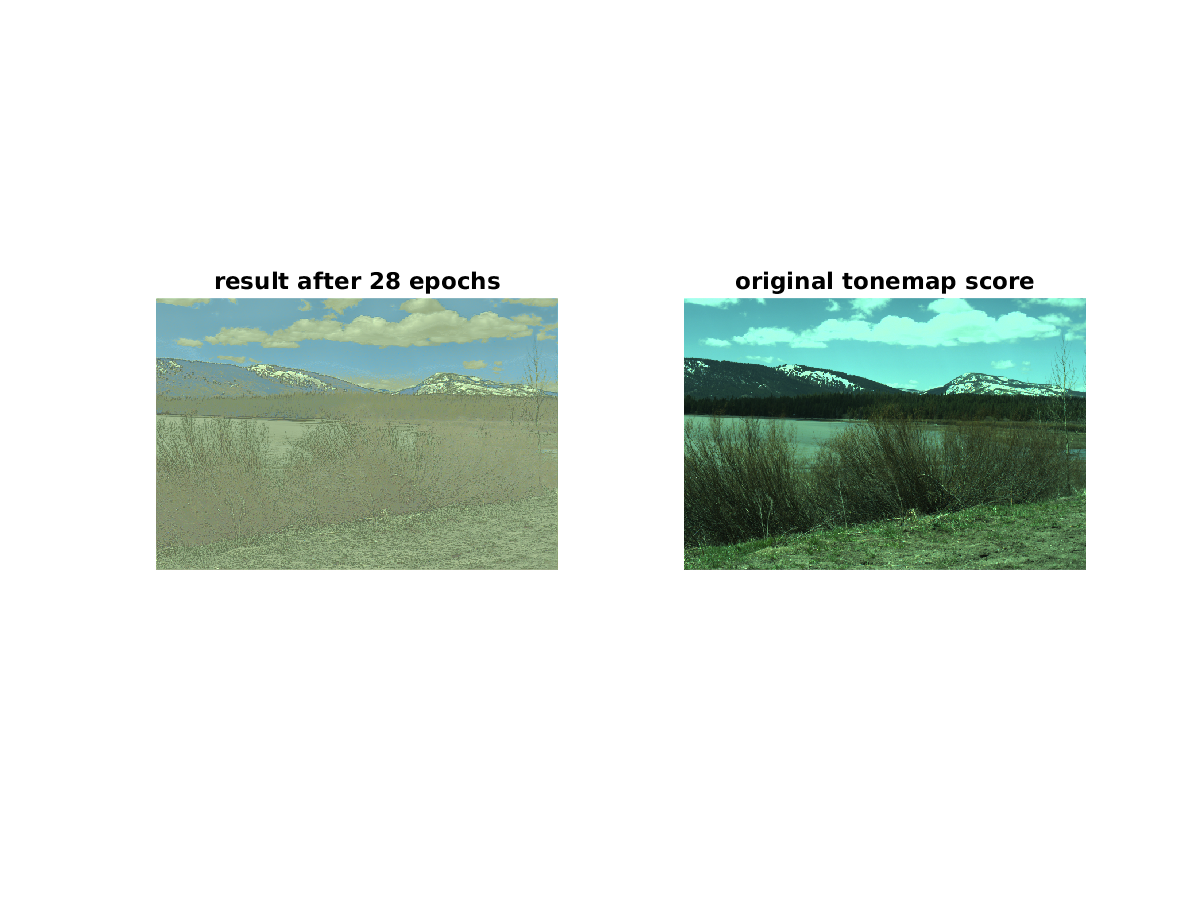}
\end{subfigure}
\caption{Some examples from test set where the tone mapping network gives plausible results.}
\end{figure}
\begin{figure}[bp!]
\begin{subfigure}{0.5\textwidth}
\includegraphics[scale = 0.5]{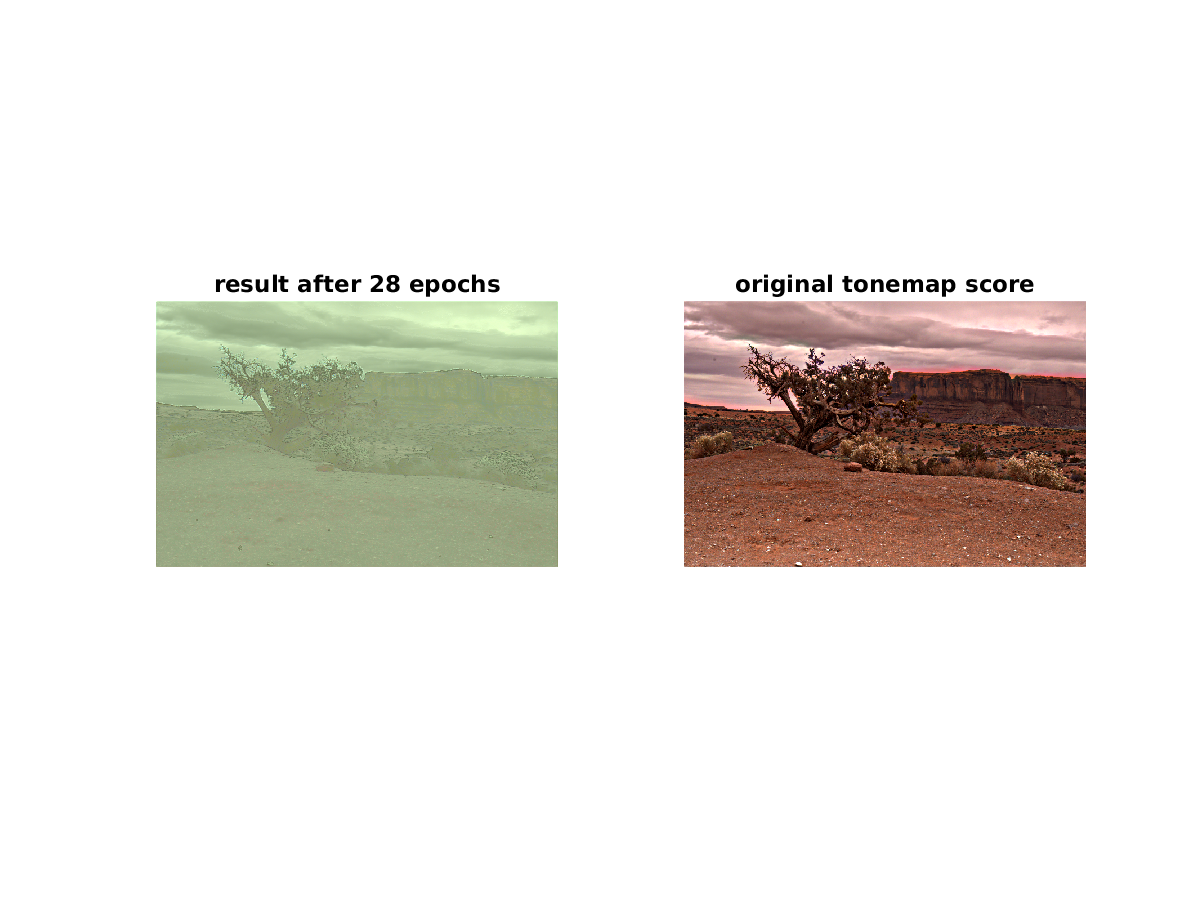}
\end{subfigure}
\begin{subfigure}{0.5\textwidth}
\includegraphics[scale = 0.5]{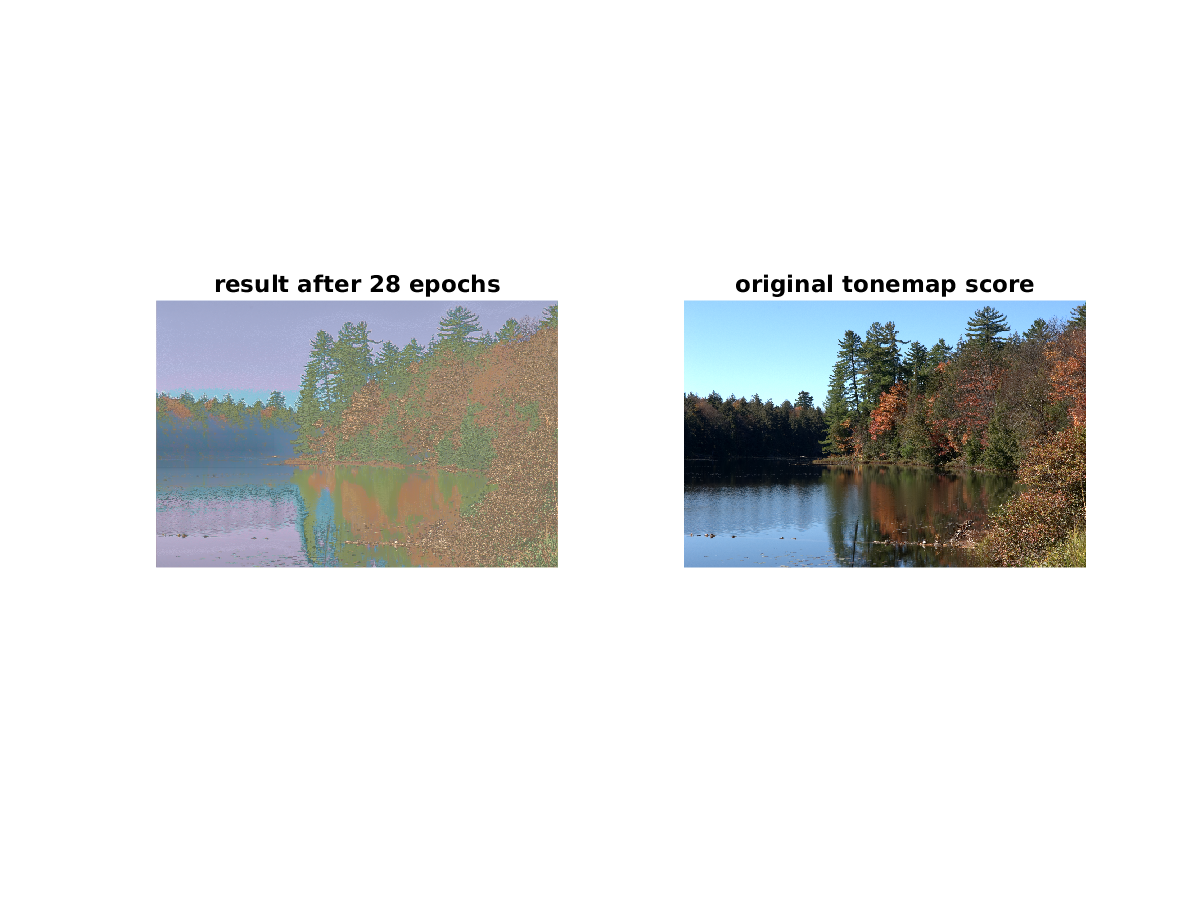}
\end{subfigure}
\caption{Some examples from test set  where the tone mapping network does not give plausible results.}
\end{figure}

\subsection{HDR2Tonemap Results}
In the results we present the outputs of the final validated architecture for the cases where the network performs good as well as bad (Fig 7. and 8.) . The final test error for that model after 28 epochs of training is 0.002764 (Fig 6.). We also present the plot of training error vs no. of epochs for that model. 
\section{References}
1.) Debevec, Paul E., and Jitendra Malik. "Recovering high dynamic range radiance maps from photographs." ACM SIGGRAPH 2008 classes. ACM, 2008.
\\
2.) Mitsunaga, Tomoo, and Shree K. Nayar. "Radiometric self calibration." Computer Vision and Pattern Recognition, 1999. IEEE Computer Society Conference on.. Vol. 1. IEEE, 1999.
\\
3.) Krizhevsky, Alex, Ilya Sutskever, and Geoffrey E. Hinton. "Imagenet classification with deep convolutional neural networks." Advances in neural information processing systems. 2012.
\\
4.) Krizhevsky, Alex, Ilya Sutskever, and Geoffrey E. Hinton. "Imagenet classification with deep convolutional neural networks." Advances in neural information processing systems. 2012.
\\
5.)Wang, Zhou, et al. "Image quality assessment: from error visibility to structural similarity." IEEE transactions on image processing 13.4 (2004): 600-612.

\end{document}